\documentstyle[aaai]{article}

\title{ACLP: Integrating Abduction\\ and Constraint Solving\thanks{This
system has been developed in collaboration with A.
 Michael and C. Mourlas. The system can be obtained from
 http://www.cs.ucy.ac.cy/aclp/.}}

\author{Antonis Kakas\\
75 Kallipoleos St.\\ CY-1678, Nicosia, Cyprus. \\
Email: antonis@cs.ucy.ac.cy}

\begin{document}

\maketitle

\begin{abstract}
\noindent
ACLP is a system which combines abductive reasoning
and constraint solving by integrating the
frameworks of Abductive Logic Programming (ALP)
and Constraint Logic Programming
(CLP). It forms a general high-level knowledge
representation environment for abductive problems in Artificial
Intelligence and other areas.
In ACLP, the task of abduction is supported and enhanced
by its non-trivial integration with constraint solving
facilitating its application to complex problems.
The ACLP system is currently implemented on
top of the CLP language of ECLiPSe as a
meta-interpreter exploiting its underlying
constraint solver for finite
domains.
It has been applied to the
problems of planning and scheduling in order to test
its computational effectiveness
compared with the direct use of the (lower level)
constraint solving framework of CLP on which
it is built. These experiments provide evidence
that the abductive framework of ACLP does not
compromise significantly the computational
efficiency of the solutions.
Other experiments show the
natural ability of ACLP to accommodate easily
and in a robust way new or changing requirements of
the original problem.
\end{abstract}

\section{Introduction}

The ACLP framework and system is an attempt to address
the problem of providing a high-level declarative
programming (or modeling) enviroment for problems of Artificial
Intelligence
which at the same time has an acceptable computational
performance. Its key elements are (i) the support of abduction
as a central inference of the system, to facilitate
a high-level of expressivity for problem representation,
and (ii) the use of constraint solving to enhance
the efficiency of the computational process of
abductive inference as this is applied on
the high-level representation of the problem
at hand.

It has been argued  in \cite{denckerkakas00} that
declarative problem solving, where the proplem representation
contains information about properties that hold true
in the problem domain rather than information on methods
of how we would solve the problem, and
abduction are closely related to each other.
In an (ideal) declarative setting
problem solving consists of filling in missing information
from the theory that represents the problem. In other words,
the solution consists of an extension of the basic description
of the problem  so that the problem
task (or goal) is satisfied in this extended description.
This process of extending the theory is called {\em abduction}.
For example, in a logical setting,
abduction as a problem solving method, assumes that the general data
structure for the solution to a problem (or solution carrier) is at
the predicate level and hence a solution is described in the same
terms and level as the problem itself.

Indeed, abduction allows a high-level representation of problems
close to their natural specification suitable for
addressing a variety of problems in AI, such as
diagnosis, planning and scheduling,
natural language understanding,
assimilation of sensor data and user modeling.
The main advantage of using abduction to solve these problems
is the high-level representation or modeling environment that
it offers. This in turn provides a high degree of
modularity and flexibility which is
useful for applications with complex and changing requirements.
But although the utility of abduction for formulating such problems in AI
is well proven there has been little work (see though
\cite{Menzies-1}) to address the question of
whether these abductive formulations can form the basis for
computationally effective solutions to realistic problems.

ACLP tries to address this problem
by a non-trivial integration of constraint solving within the
abductive process.
The general pattern of computation in ACLP
consists of a cooperative
interleaving between hypotheses and constraint generation, via
abductive inference,
with
constraint satisfaction of the generated constraints.
Abductive reasoning provides an incremental
reduction of the high-level problem representation and
goals to abductive hypotheses together with lower-level constraints
whose form is problem independent.
The integration of abductive reasoning with constraint
solving in ACLP is cooperative, in the sense that the
constraint solver not only solves the final constraint store
generated by the abductive reduction but also affects
dynamically this abductive search for a solution. It enables
abductive reductions to be pruned early by setting new
suitable constraints on the abducible assumptions into the
constraint store, provided that this remains satisfiable.
During the ACLP computation there is a non-trivial
interaction between (i) reduction of goals and consistency
checking of abducible assumptions, (ii) setting new
constraints in the constraint store of reduction and (iii)
generating further abductive hypotheses.

\section{General Information}

Currently, the ACLP system is implemented as a meta-interpreter
on top of the CLP language of ECLiPSe. As such the
system is relatively compact comprising about 500 lines of code.
It is based on the abductive proof procedure developed in
\cite{KM-ICLP95} (which in turn
follows a series of proof procedures
\cite{EK-89}, \cite{KM-VLDB90}, \cite{KM-PRICAI90})
and uses the CLP constraint solver of ECLiPSe to handle
constraints over finite domains (integer and atomic elements).
The architecture of the system is quite general and can be
implemented in a similar way with other constraint solvers.

The ACLP system runs on any platform on which ECLiPSe runs. This
includes all major platforms.
It can be obtained, together with information on how to use it,
from the following web address: http://www.cs.ucy.ac.cy/aclp/.
ACLP programs (see section~\ref{applyingACLP} below) are loaded
into ECLiPSe together with the ACLP system file, {\tt aclp.pl},
and executed
by calling the top-level ECLiPSe query: \\
\centerline{\em aclp-solve(+Goal, +Initial-hypothesis, ?output-variable).}
The output-variable returns a list of abducible hypotheses, with their
domain variables constrained according to the dynamic constraints that were
generated through the unfolding of the ``relevant'' part, with
respect to the Goal, of the
program and the integrity constraints.
A subsequent step of {\em labelling}  on these
variables is needed to give a ground solution of our query, +Goal.
Various constraint predicates of ECLiPSe can be used at this stage
e.g. {\em min\_max/2} or {\em minimize/2} to find an optimal
ground solution. If we are not interested in such further optimization
we can use the simpler queries:\\
\centerline{\em aclp-solve(+Goal)}
\centerline{\em aclp-solve(+Goal, +Initial-hypothesis).}

The initial-hypothesis variable is a list of ground abducible facts
which we want the system to take as given when constructing a solution.
It is used when we have partial information about the solution
that we are looking for.
(If no such information is known then this is given as the empty list.)
Its typical use is when we want
to {\em recompute} the solution to a goal under
some new requirements by adapting the old solution as for example in the
case of rescheduling.
The old solution (or part of this)
will then form the initial-hypothesis.



\section{Applying the System}
\label{applyingACLP}


The ACLP system is a programming environment on top of
the ECLiPSe language. An ACLP program is an abductive
theory consisting  of a triple $<P,A,IC>$ where:
\begin{itemize}

\item $P$ is a finite set of user-defined ECLiPSe  clauses,

\item $A$ is
a set of declarations of abducibles
predicates in the form of ECLiPSe facts as:
$abducible\_predicate(predicate\_name/arity)$,

\item $IC$ is a set of integrity constraints written as ECLiPSe
rules of the form: $ ic :- B_1,\ldots,B_n. \ \ \ (n \ge 1),$
where:
\begin{itemize}
\item at least one of the goals $B_1,\ldots,B_n$ has
      an abducible predicate, and
\item the rest of the goals can be either positive or negative
literals on user-defined predicates or constraint predicates of ECLiPSe.
\end{itemize}
\end{itemize}

The (lower-level) problem independent CLP
constraint predicates that can be used in the body
of a program rule or an integrity constraint
can be (i) {\em arithmetic constraint
predicates (over the integers)}
or (ii) {\em logical contraint predicates}.
The constraint predicates on finite domain variables
of ECLiPSe that are
supported by the current ACLP implementation are:

\begin{description}
\item[{\tt T1\#\#T2\ :}] the value of variable T1 is not equal to that
of variable T2.
\item[{\tt T1\#=T2\ :}] the value of variable T1 is equal to that of
variable T2.
\item[{\tt T1\#<T2\ :}] the value of variable T1 is less than that of
variable T2.
\item[{\tt T1\#<=T2:}] the value of variable T1 is less or equal to
that of variable T2. 
\item[{\tt T1\#>T2\ :}] the value of variable T1 is greater than that
of variable T2.
\item[{\tt T1\#>=T2:}] the value of variable T1 is greater or equal
to that of variable T2. 
\end{description}

The equality and inequality constraints are also supported over
other non-arithmetic user-defined finite domains. The system
also has a term equality constraint, {\tt T1\#\#=T2\ },
where the terms $T1$ and $T2$ can contain variables one level deep
inside a function sympol.
In addition, ACLP supports logical constraints such as
conjunction, {\tt \#$\land$}, and disjunction, {\tt \#$\vee$}. These simple
constraints can be combined to build
complex logical constraint expressions.
During the ACLP computation constraints maybe negated
and their negation is set in the current constraint store.
This negation of the constraints is the usual mathematical negation, e.g.
the negation of the aritmetic constraint {\tt T1\#<T2} is {\tt T1\#>=T2}.
The negation of the logical constraint {\tt \#$\land$} is
{\tt \#$\vee$}.

An ACLP program, $<P,A,IC>$, can contain negation as failure literals
in $P$ and $IC$.
Negation as failure is handled through abduction simply as
another type of abducible in the theory.
All occurences of $not(p)$ in
the program $P$ are replaced by $not\_p$ which is treated
as an abducible with the canonical integrity
constraint $ic:- \ not\_p, \ p.$ In the current
implementation it is necessary for the user to
specify explicitly both the fact that $not\_p$ is
abducible by adding a statement
$abducible\_predicate(not\_p/arity)$ in the program as well
as adding the above canonical constraint
in the program.
The semantics of negation as failure is that of (partial) Stable
Models in the program $P$  and Generalised (partial) Stable Models
when we consider the whole abductive theory with its integrity
constraints $IC$. The details of the abductive semantics for
ACLP programs and the particular treatment of NAF can be found
in \cite{KMM-aclp00}.

As an example, the ACLP program below is an
implementation of the basic axioms (of persistence)
of the Event Calculus \cite{Sergot-EC}
suitable for abductive planning. The program $P$ consists of
the following clauses:
\begin{verbatim}
   holds_at(P,E) :- initially(P,T),
                    not clipped(T,E,P).
   holds_at(P,E) :- initiates(P,A),
                    time(T), T #< E,
                    act(T,A),
                    not clipped(T,E,P).
\end{verbatim}
\noindent
together with the auxiliary definitions:
\begin{verbatim}
   between(A,B,C) :- A #<B, B#<C.
   time(T) :- maximum_time(Max), T :: 1..Max.
\end{verbatim}

\noindent
The abducible predicates in $A$ are the action predicate
$act/2$ and the NAF predicate $not\_clipped/3$ declared by the
following clauses:
\begin{verbatim}
   abducible_predicate(act/2).
   abducible_predicate(not_clipped/3).
\end{verbatim}
\noindent
The integrity constraints in $IC$ contain the negation
as failure constraint as the clause:
\begin{verbatim}
   ic :- not_clipped(T,E,P), terminates(P,A1),
         act(C,A2), A1 ##= A2, between(T,C,E).
\end{verbatim}
\noindent
and contraints that encode the preconditions of actions written
as clauses of the general form:
\begin{verbatim}
   ic :- act(T,A), not preconditions(A,T).
\end{verbatim}
\noindent
In a specific planning domain, e.g. the trucks domain, this
will be extended with clauses for the $initiates$ and $terminates$
predicates in $P$, for example:
\begin{verbatim}
   initiates(in(Obj,Truck),
             load_truck(Obj,Truck,Loc)).
   terminates(at(Obj,Loc),
             load_truck(Obj,Truck,Loc)).
\end{verbatim}
\noindent
and the definitions, again in $P$, of the preconditions
of the specific actions in the domain, for example:
\begin{verbatim}
   preconditions(load_truck(Obj,Truck,Loc),T):-
         holds_at(at(Obj,Loc),T),
         holds_at(at(Truck,Loc),T).
\end{verbatim}
\noindent
The initial state is defined by a set of facts of the
form $initially(Property,0)$, e.g.  $initially(at(package1,city1\_1),0).$

\subsection{Methodology}

  %
ACLP has been applied to several different types of abductive problems
such as planning, air-crew scheduling, optical music recognition,
analysis of software requirements and intelligent information integration
(see below section\ref{users}).
Although most of these applications
are not of "industrial scale" they
indicate some methodological guidelines
that can be followed when using ACLP. As ACLP is a
general development framework with no
specific application domain these
guidelines can only be themselves of a
general nature.

The central advantage of an abductive approach
is the high-level declarative representation that it allows.
This means that the development of the program can be
done incrementally starting first with a "pure"
declarative representation based on a simple model
of the problem and gradually refine this model to reflect
more and more particular domain knowledge of the problem at hand.
A central first decision to be taken is the choice of
abducibles for the problem. These play the important role
of the solution carriers or answers to the problem
goals.
Each problem has its own abducible answer predicates
(c.f. the usual answer holder of a logical variable
in LP and CLP ) which means that we can describe
directly in our theory (the ACLP program)
the desired properties of the solution.

An important methodological step is the
distinction between strict validity requirements
on the solution of our problem, which are separated in the
integrity constraints $IC$ of the ACLP theory,
and the basic model of our problem which is described
in the program $P$ of the ACLP theory. A good such
separation means that we can then
incrementally refine this basic model to improve the
quality of the solution without affecting its validity
(which is always ensured by the integrity constraints in
$IC$). As we refine our representation we include
more domain specific information, exploiting
any natural structures of the problem at hand,
that can help to improve the computation of the
solution.

At a final step of refinement of the
problem representation we can develop the model
in order to control the choice of abducible
in the abductive reduction of the problem goals.
This choice can be implemented to follow either some heuristics,
priorities, or algorithm for optimality, to
control both the computational efficiency
and the quality of the solutions.
We can then experiment
with different design alternatives adopting different
strategies to study how this would affect the quality of the solutions.
In large scale problems the user can also experiment with
different orders in which the integrity constraints are
satisfied. Generally, the heuristic of trying first more
specific integrity constraints gives better results.

An importnat characteristic
of an ACLP representation of a problem is the
flexibility it offers under new or dynamically changing
requirements. Once we have one complete
representation of the problem we can easily experiment
with different requirements on the solution, by changing
the integrity constraints which specialize the general
model to the needs and preferences of a particular case.
This can be done in a modular way by affecting only the
integrity constraints to reflect the new requirements.


\subsection{Users and Useability}
\label{users}

ACLP is a high-level knowledge representation environment
which supports directly abduction. Its use requires some basic
knowledge of logic programming, constraint logic programming
\cite{CLP-survey-98} and abductive logic programming \cite{KKT-survey}.
As it is implemented on top
of ECLiPSe knowledge of this particular CLP language can help.
In some cases it is also useful to understand some of the basic search
heuristics that ACLP and ECLiPSe underneath use in their
computation (see below in section \ref{evaluation}).
Details of how to use it with examples can be found at the
web page of ACLP at: http://www.cs.ucy.ac.cy/aclp/.

The ACLP framework as a declarative problem solving
paradigm can be used to address several different
types of problems. Its developers have applied it
initially to the problems of scheduling and planning
\cite{KM-LPNMR97,KMM-nmr98}
to test its computional effectiveness and its flexibility
in problem representation. Also it has been used
in an industrial application of crew-schedulling \cite{KM-IEA99}.
Other groups have used ACLP for (i) optical music recognition
\cite{aclpmusic} where ACLP was used to implement a system
that can handle recognition under incomplete information,
 and (ii) resolving inconsistencies in
software requirements \cite{russo99}
where (a simplified form of) ACLP was used to identify the
causes of inconsistency and suggest changes that can restore
consistency of the specification. Also the intelligent information
integration work of \cite{coin} although it does not use
ACLP in its implementation its approach to information integration
is based on an ACLP representation.

Currently, we are considering two new applications of ACLP.
One is that of the development of an information integration
mediator for integrating information suitable for electronic
commerce applications. The other application area concerns
the further development of the problem of planning with
emphasis on (i) the study of a systematic way to exploit domain
specific information,  and (ii) the problem of planning
under incomplete information about the initial state of the
problem.


We also mention that ACLP programs can be generated automatically
from example data using a machine learning technique called
{\em abductive concept learning}. For details of this method
and a related system see http://www-lia.deis.unibo.it/Software/ACL/.

\section{Evaluating the System}
\label{evaluation}

 %
 %
 %
 %

At this initial stage of the development of the ACLP system the
main aim of its evaluation is to understand the cost of the extra
high-level expressivity layer that it gives (over for example CLP
approaches) in comparison with the advantages of
modularity that this may provide.
The ACLP system has thus been evaluated mainly in two different directions:
(i) computational efficiency, particularly in comparison with the
underlying CLP language of ECLiPSe on which it is implemented, and
(ii) flexibility under changes of the problem specification.
The overall evaluation of an application under ACLP is a combination
of these two factors together with the quality of the generated
solutions under some optimization criteria when such criteria
apply.

For example, the air-crew scheduling application in
\cite{KM-IEA99} produced solutions (for the small sized company of
Cyprus Airways) that were judged to be of good quality,
comparable to manually generated solutions by experts of many
years on the particular problem, while at the same time
it provided a flexible platform on which the company\footnote{
Unfortunately, the company has decided not to use the system for
reasons that relate more to their policy of adopting a global
solution to the full computerization of their operation with
direct compatibility between their different systems.}
could easily experiment with changes in policy and preferences. Also
the re-scheduling module of the system was judged to be of
high-value both as a tool for adjusting the initially generated
solution and for handling unexpected changes on the day
of operation.

The computational effectiveness of the ACLP
system depends on two factors:
(a) the effectiveness of the reduction of the high-level
ACLP representation to lower-level finite domain constraints
and (b) the effeciency of the underlying constraint solver
in propagating (or solving) these constraints. In fact, these
two factors are interrelated as in many cases
the reduction in (a) depends on the completeness of the
propagation in (b). For some problems where these are
not strongly related e.g. in the case of
job-shop sheduling we can see that in comparison
with (b) the overhead for the reduction in (a) is small.
Information on these evaluation experimenents can be found at the
ACLP web pages.
Further results of comparison on problems where
factors (a) and (b) are loosesly coupled can be found in the
recent work of \cite{pelov00} where
experiments with various
types of systems, including ACLP, on the constraint satisfaction
problems of the N-queens and graph colouring have been perfomed.

Another, but limited, comparison that we have carried out in order
to test the effectiveness of the current ACLP implementation
was a comparison with the use of Constraint Handling Rules (CHRs)
\cite{CHR} on the same problems of job-shop scheduling. On the
whole the ACLP system was at least as effecient as CHR.
It should be noted though that these comparions were carried
out before recent developments on CHRs.

In problems where the search space of the reduction
of the high-level specification depends
strongly on the fast detection that the contraint store of
finite domain constraints is becoming unsatisfiable, the
overall computational efficiency of ACLP can be sensitive
to the particular way of modeling the problem and the
amount of specific domain knowledge it contains
pertaining to the computation aspects of the problem.
Such a problem is that of planning.
As an example, table~\ref{table1} shows the
execution time (on a SUN Ultra-1 with
64Mb RAM) and the number of moves
required for an indicative set of blocks world planning
problems. The representation of the problem that was used
was purely declarative with the exception
that the towers of the final state were to be build
in a horizontal fashion from the bottom up. The number
of available positions on the table was restricted to be
one third of the total number of blocks in order to make the
problems more computationally demanding.
These times are comparable with the execution times of solving such
problems directly in ECLiPSe as reported in \cite{ICPlan}.

\begin{table}[t]
\begin{center}
\begin{tabular}{|c|c|c|}  \hline
\multicolumn{3}{|c|}{\bf Performance Measurements} \\ \hline
 Blocks   &  time (in secs) & \# of moves  \\ \hline
 15     & 3.7    & 23    \\ \hline
 24     & 15.19  & 34     \\ \hline
 33     & 40.69  & 49    \\ \hline
 42     & 69.5   & 58    \\ \hline
 51     & 325.97 & 74    \\ \hline
 60     & 345.52 & 87    \\ \hline
 69     & 662.15 & 101 \\ \hline
\end{tabular}
\vspace*{0.2cm}
\caption{Performance on Blocks World Planning}
\label{table1}
\end{center}
\end{table}

The flexibility of the ACLP system as a knowledge representation
framework is tested by examining how easy it is for a given ACLP
representation to be adapted under changes in the requirements
of the original problem. There are two factors to measure here:
(1) the programming effort required to adapt an existing
solution to the new problem,
and (2) the computational robustness of the system under
such changes. Experiments in the domains
of job-shop scheduling, air-crew scheduling
and planning show that the extra programming
effort in ACLP is considerably
smaller than the corresponding effort when the problem is
represented directly in ECLiPSe. In many cases, the
effort required in ACLP is simply the addition of some
new integrity constraints written directly from the
declarative specification of the new requirements.
The same experiments show
that on the whole the computational performance of ACLP
remains within the same order of magnitute under
changes which affect the problem only locally in one part,
e.g. extra requirements on the moves allowed for
a particular "small" subset of blocks in
the problem of blocks world planning.

Another feature of the flexibility of ACLP is the ability to
use it to recompute the solution for a given goal, under some
new information about the particular instance of the problem,
so that the new solution remains "close" to the old solution,
e.g. it contains a minimal number of changes from the old solution.
Experiments to test this feature have been performed
on the problems of job-shop and air-crew scheduling.
Table~\ref{resched-table} shows an example of
the results.
For each one of these problems
a new requirement of some resource unavailability
was added (shown below in the first column).
The rescheduling results are
shown in the third column of the table, which gives
the time together with the number of
changes needed on the existing solution in order to
satisfy the new requirements. The fourth column displays
the analogous information for the control
experiment of re-executing the goal with the extra requirement
represented in the program but now without any initial solution.

\begin{table*}[t]
\begin{center}
\begin{tabular}{|c||c|c|c|}  \hline
\multicolumn{4}{|c|}{\bf Performance \& Quality Measurements }
\\ \hline
Problem   &  Initial Solution & Rescheduling & Re-Execution  \\
\# tasks - Resource Availability & $secs$  & $secs$/$changes$ &
   $secs$/$changes$ \\ \hline
25 tasks  &                   &              &               \\
$R4$ unavailable in period (0,20) & 0.23 & 0.27/4 & 0.23/21 \\ \hline
50 tasks  &                   &              &               \\
$R2$ unavailable in period (20,30) & 0.71 & 1.94/14 & 0.74/40 \\ \hline
75 tasks  &                   &              &               \\
$R4$ unavailable in period (0,16) &  1.03 & 1.25/5  & 1.81/25 \\ \hline
100 tasks  &                   &              &               \\
$R7$ unavailable in period (20,35) & 1.61 & 1.95/1  & 1.67/47 \\ \hline
\end{tabular}
\vspace*{0.2cm}
\caption{Rescheduling Experiments for Resource Unavailability in
Job-shop Scheduling}
\label{resched-table}
\end{center}
\end{table*}

\subsection{Future Development}

ACLP is a general purpose declarative programming framework.
Hence its evaluation must combine different aspects
of its performance. At this initial stage the emphasis in the
development of the first prototype was on its declarativeness
together with an acceptable computational performance.

The search that ACLP performs in constructing a solution
needs further study for improvement. Currently, the system
employs a few simple heuristics to help in its search.
As ACLP is parametric on the underlying finite domain
constraint solver improvements on its performance will
improve the ACLP performance. More important though,
is the interaction between the abductive reduction and
the satisfaction of the finite domain constaints that it generates.
This is the major aspect of the search space of ACLP.
Hence one way to improve the ACLP search is to develop further the
interface between the abductive reduction
and the constraint solver so that the propagation of the
domain constraints varies
according to some heuristic criteria
on the point of the search space
where the request to the constraint solver is made.

In particular, while the abductive
process is reducing the high-level goal and integrity
constraints there are choice points where we can
either introduce a new abducible hypothesis in the
solution or instead backtrack higher up in the search
space. In many cases, this decision can be made to
depend on the satisfaction of some of the lower-level domain
constraints that are generated by the abductive reduction.
We can then evaluate the significance of their
satisfaction (currently the system adopts a very simple form
of evaluation) and depending on this guide the
search to introduce or not a
new hypotheses in the solution.
This has to be combined
with the general heuristic of abductive search of prefering to
reuse hypotheses and delay the specialisation of (non-ground)
hypotheses or the generation of new ones.
In developing though a better search for ACLP it maybe necessary to restrict
our attention to separate classes of problems e.g. to develop
separately an ACLP planner from an ACLP system for diagnosis.

These considerations of improving the general purpose
search strategy of the system is an important
next stage of development. On the other hand,
it is clear that the general improvement of
efficiency that can be achieved is limited as we are aiming
to use the system for computational hard problems.
Hence another line of development is to provide
more facilities for problem specific information to be
incorporated in the representation of the problem whose
exploitation can improve the performance of the system
on the particular problem at hand. This problem
specific information could include information to
directly control the search of the system on the
particular problem in the same spirit of recent developments
for controlling models in constraint programming \cite{OPL}.


\small

\bibliography{NMR2000-aclp-cam}

\begin{thebibliography}{}

\bibitem[\protect\citeauthoryear{Bressan \& Goh.}{1997}]{coin}
Bressan, S., and Goh., C.
\newblock 1997.
\newblock Semantic integration of disparate information sources over the
  internet using constraints.
\newblock In {\em Constraint Programming Workshop on Constraints and the
  Internet}.

\bibitem[\protect\citeauthoryear{Denecker \& Kakas}{2000}]{denckerkakas00}
Denecker, M., and Kakas, A.~C.
\newblock 2000.
\newblock Abductive logic programming: Editorial forward.
\newblock {\em Journal of Logic Programming: Special Issue of Abductive Logic
  Programming}.

\bibitem[\protect\citeauthoryear{El-Kholy \& Richards}{1996}]{ICPlan}
El-Kholy, A., and Richards, B.
\newblock 1996.
\newblock Temporal and resource reasoning in planning: the parcplan approach.
\newblock In {\em Proceedings of ECAI-96}.

\bibitem[\protect\citeauthoryear{Eshghi \& Kowalski}{1989}]{EK-89}
Eshghi, K., and Kowalski, R.
\newblock 1989.
\newblock Abduction compared with {N}egation by {F}ailure.
\newblock In {\em Proceedings of the 6th International Conference on Logic
  Programming}.

\bibitem[\protect\citeauthoryear{Ferrand, Leite, \& Cardoso}{1999}]{aclpmusic}
Ferrand, M.; Leite, J.; and Cardoso, A.
\newblock 1999.
\newblock Improving optical music recognition by means of abductive constraint
  logic programming.
\newblock In {\em Proceedings of EPIA'99}, number 1695 in LNAI,  342--356.

\bibitem[\protect\citeauthoryear{Fruhwirth}{1998}]{CHR}
Fruhwirth, T.
\newblock 1998.
\newblock Theory and practice of constraint handling rules.
\newblock {\em Journal of Logic Programming: Special Issue on Constraint Logic
  Programming} 37(1-3).

\bibitem[\protect\citeauthoryear{Hentenryck}{1999}]{OPL}
Hentenryck, P.~V.
\newblock 1999.
\newblock {\em Optimization Programming Language}.
\newblock MIT Press.

\bibitem[\protect\citeauthoryear{J.Jaffar \& M.J.Maher}{1998}]{CLP-survey-98}
J.Jaffar, and M.J.Maher.
\newblock 1998.
\newblock Constraint logic programming: A survey.
\newblock In Gabbay, D.; Hogger, C.; and Robinson, J., eds., {\em Handbook of
  Logic in AI and Logic Programming}, volume~5. Oxford University Press.
\newblock  591--696.

\bibitem[\protect\citeauthoryear{Kakas \& Mancarella}{1990a}]{KM-VLDB90}
Kakas, A., and Mancarella, P.
\newblock 1990a.
\newblock Database updates through abduction.
\newblock In D.~McLeod, R. S.-D., and Schek, H., eds., {\em Proceedings of the
  16th International Conference on Very Large Databases, VLDB-90},  650--661.
\newblock Morgan Kaufmann.

\bibitem[\protect\citeauthoryear{Kakas \& Mancarella}{1990b}]{KM-PRICAI90}
Kakas, A., and Mancarella, P.
\newblock 1990b.
\newblock On the relation between truth maintenance and abduction.
\newblock In {\em Proceedings of the 2nd Pacific Rim International Conference
  on Artificial Intelligence}.

\bibitem[\protect\citeauthoryear{Kakas \& Michael}{1995}]{KM-ICLP95}
Kakas, A., and Michael, A.
\newblock 1995.
\newblock Integrating abductive and constraint logic programming.
\newblock In {\em Procedings of the 12th International Conference on Logic
  Programming, ICLP95}.

\bibitem[\protect\citeauthoryear{Kakas \& Michael}{1999}]{KM-IEA99}
Kakas, A., and Michael, A.
\newblock 1999.
\newblock Air-crew scheduling through abduction.
\newblock In {\em Proceedings of IEA/AIE-99},  600--612.

\bibitem[\protect\citeauthoryear{Kakas \& Mourlas}{1997}]{KM-LPNMR97}
Kakas, A., and Mourlas, C.
\newblock 1997.
\newblock Aclp: Flexible solutions to complex problems.
\newblock In {\em Proceedings of Logic Programming and Non-monotonic Reasoning,
  LPNMR97}.

\bibitem[\protect\citeauthoryear{Kakas, Kowalski, \& Toni}{1998}]{KKT-survey}
Kakas, A.; Kowalski, R.; and Toni, F.
\newblock 1998.
\newblock The role of abduction in logic programming.
\newblock In Gabbay, D.; Hogger, C.; and Robinson, J., eds., {\em Handbook of
  Logic in AI and Logic Programming}, volume~5. Oxford University Press.
\newblock  235--324.

\bibitem[\protect\citeauthoryear{Kakas, Michael, \& Mourlas}{1998}]{KMM-nmr98}
Kakas, A.; Michael, A.; and Mourlas, C.
\newblock 1998.
\newblock Aclp: a case for non-monotonic reasoning.
\newblock In {\em Proceedings of NMR98},  46--56.

\bibitem[\protect\citeauthoryear{Kakas, Michael, \& Mourlas}{2000}]{KMM-aclp00}
Kakas, A.; Michael, A.; and Mourlas, C.
\newblock 2000.
\newblock Aclp: Abductive constraint logic programming.
\newblock {\em Journal of Logic Programming: Special Issue of Abductive Logic
  Programming}.

\bibitem[\protect\citeauthoryear{Kowalski \& Sergot}{1986}]{Sergot-EC}
Kowalski, R., and Sergot, M.
\newblock 1986.
\newblock A logic-based calculus of events.
\newblock {\em Journal of New Generation Computing} 4.

\bibitem[\protect\citeauthoryear{Menzies}{1996}]{Menzies-1}
Menzies, T.
\newblock 1996.
\newblock Applications of abduction: Knowledge-level modeling.
\newblock {\em Journal of Human Computer Studies}.

\bibitem[\protect\citeauthoryear{Pelov, Mot, \& Bruynooghe}{2000}]{pelov00}
Pelov, N.; Mot, E.~D.; and Bruynooghe, M.
\newblock 2000.
\newblock A comparison of logic programming approaches for representation and
  solving of constraint satisfaction problems.
\newblock In {\em Proceedings of NMR2000: special session on Abduction}.

\bibitem[\protect\citeauthoryear{Russo \bgroup \em et al.\egroup
  }{1999}]{russo99}
Russo, A.; Miller, R.; Nuseibeh, B.; and Kramer, J.
\newblock 1999.
\newblock An abductive approach for handling inconsistencies in scr
  specifications.
\newblock Technical report, Imperial College.

\end{thebibliography}
\bibliographystyle{aaai}

\end{document}